\documentclass{article}

\usepackage[final]{neurips_2018}
\usepackage{natbib}
\setcitestyle{round}

\usepackage[utf8]{inputenc} 
\usepackage[T1]{fontenc}    
\usepackage[draft]{hyperref}       
\usepackage{url}            
\usepackage{booktabs}       
\usepackage{amsfonts}       
\usepackage{nicefrac}       
\usepackage{microtype}      
\usepackage{amsmath}
\usepackage{amssymb}
\usepackage{graphicx}
\usepackage{caption}
\usepackage{algorithm}
\usepackage{algpseudocode}
\usepackage{array}
\newcolumntype{L}[1]{>{\raggedright\let\newline\\\arraybackslash\hspace{0pt}}m{#1}}

\usepackage{color}
\newcommand{\ignore}[1]{}

\algnewcommand\algorithmicinput{\textbf{Input:}}
\algnewcommand\algorithmicoutput{\textbf{Output:}}
\algnewcommand\Input{\item[\algorithmicinput]}%
\algnewcommand\Output{\item[\algorithmicoutput]}%

\newcommand{\B}[1]{\mathbf{#1}}

\title{Unsupervised Text Style Transfer using Language Models as Discriminators}

\author{
  Zichao Yang$^1$, Zhiting Hu$^1$, Chris Dyer$^2$, Eric P. Xing$^1$, Taylor Berg-Kirkpatrick$^1$\\
  $^1$Carnegie Mellon University, $^2$DeepMind\\
  \texttt{\{zichaoy, zhitingh, epxing, tberg\}@cs.cmu.edu} \\
  \texttt{cdyer@google.com} \\
}

\begin{document}

\maketitle

\begin{abstract}
  Binary classifiers are often employed as discriminators in GAN-based
  unsupervised style transfer systems to ensure that transferred sentences are
  similar to sentences in the target domain. One difficulty with this approach
  is that the error signal provided by the discriminator can be unstable and is
  sometimes insufficient to train the generator to produce fluent language. In
  this paper, we propose a new technique that uses a target domain language
  model as the discriminator, providing richer and more stable token-level
  feedback during the learning process.
  We train the generator to minimize the negative log likelihood (NLL) of
  generated sentences, evaluated by the language model. By using a continuous
  approximation of discrete sampling under the generator, our model can be
  trained using back-propagation in an end-to-end fashion. Moreover, our
  empirical results show that when using a language model as a structured
  discriminator, it is possible to forgo adversarial steps during training,
  making the process more stable. We compare our model with previous work that uses
  convolutional networks (CNNs) as discriminators, as well as a broad set of other approaches. Results show that the proposed method achieves improved performance on three tasks: word substitution
  decipherment, sentiment modification, and related language translation.
\end{abstract}
\medskip

\section{Introduction}
Recently there has been growing interest in designing natural language
generation (NLG) systems that allow for control over various attributes of
generated text -- for example, sentiment and other stylistic properties. Such
controllable NLG models have wide applications in dialogues
systems~\citep{wen2016network} and other natural language interfaces. Recent
successes for neural text generation models in machine
translation~\citep{bahdanau2014neural}, image captioning~\citep{vinyals2015show}
and dialogue~\citep{vinyals2015neural,wen2016network} have relied on massive
parallel data. However, for many other domains,
only non-parallel data -- which includes collections of sentences from each
domain without explicit correspondence -- is available. Many text style transfer
problems fall into this category. The goal for these tasks is to transfer a
sentence with one attribute to a sentence with an another attribute, but with the
same style-independent content, trained using only non-parallel data.

Unsupervised text style transfer requires learning disentangled representations
of attributes (e.g., negative/positive sentiment, plaintext/ciphertext orthography)
and underlying content.
This is challenging because the two interact in subtle ways in natural language
and it can even be hard to disentangle them with parallel data. The recent
development of deep generative models like variational auto-encoders
(VAEs)~\citep{kingma2013auto} and generative adversarial networks(GANs)~\citep{goodfellow2014generative} have made learning disentangled
representations from non-parallel data possible. However, despite their rapid
progress in computer vision---for example, generating photo-realistic
images~\citep{radford2015unsupervised}, learning interpretable
representations~\citep{chen2016infogan}, and translating
images~\citep{zhu2017unpaired}---their progress on text has been more limited.
For VAEs, the problem of training collapse can severely limit
effectiveness~\citep{bowman2015generating,yang2017improved}, and
when applying adversarial training to natural language, the
non-differentiability of discrete word tokens makes generator optimization
difficult. Hence, most attempts use REINFORCE~\citep{sutton2000policy} to
finetune trained models~\citep{yu2017seqgan, li2017adversarial} or uses
professor forcing~\citep{lamb2016professor} to match hidden states of decoders.

Previous work on unsupervised text style
transfer~\citep{hu2017toward,shen2017style} adopts an encoder-decoder
architecture with style discriminators to
learn disentangled representations. The encoder takes a sentence as an
input and outputs a style-independent content representation. The
style-dependent decoder takes the content representation and a style
representation and generates the transferred sentence. 
\cite{hu2017toward} use a style classifier to directly enforce 
the desired style in the generated text. \cite{shen2017style} 
leverage an adversarial training scheme where a binary CNN-based 
discriminator is used to evaluate whether a transferred sentence is real or fake,
ensuring that transferred sentences match real sentences in terms of target
style. However, in practice, the error signal from a binary classifier is
sometimes insufficient to train the generator to produce fluent language, and
optimization can be unstable as a result of the adversarial training step.

We propose to use an implicitly trained
language model as a new type of discriminator, replacing the more conventional
binary classifier.
The language model calculates a sentence's likelihood, which decomposes into a
product of token-level conditional probabilities.
In our approach, rather than training a binary classifier to distinguish real
and fake sentences, we train the language model to assign a high probability to
real sentences and train the generator to produce sentences with high
probability under the language model.
Because the language model scores sentences directly using a product of locally
normalized probabilities, it may offer more stable and more useful training
signal to the generator. Further, by using a continuous approximation of
discrete sampling under the generator, our model can be trained using
back-propagation in an end-to-end fashion.

We find empirically that when using the language model as a structured
discriminator, it is possible to eliminate adversarial training steps that use
negative samples---a critical part of traditional adversarial training.
Language models are \emph{implicitly} trained to assign a low probability to negative
samples because of its normalization constant. By eliminating the adversarial
training step, we found the training becomes more stable in practice.

To demonstrate the effectiveness of our new approach, we conduct experiments on
three tasks: word substitution decipherment, sentiment modification, and related
language translation. We show that our approach, which uses only a language
model as the discriminator, outperforms a broad set of state-of-the-art approaches on the three tasks.

\section{Unsupervised Text Style Transfer}
We start by reviewing the current approaches for unsupervised
text style transfer~\citep{hu2017toward,shen2017style}, and then
go on to describe our approach in Section~\ref{sec:lm}. Assume we have two text
datasets $\B{X} = \{\B{x}^{(1)}, \B{x}^{(2)}, \ldots, \B{x}^{(m)}\}$ and $\B{Y} =
\{\B{y}^{(1)}, \B{y}^{(2)}, \ldots, \B{y}^{(n)} \}$  with two different styles $\B{v}_x$ and $\B{v}_y$, respectively. For example, $\B{v}_x$ can be the positive sentiment style and $\B{v}_y$ can be the negative sentiment style. The datasets are
non-parallel such that the data does not contain pairs of $(\B{x}^{(i)},
\B{y}^{(j)})$ that describe the same content. The goal of style transfer is to
transfer data $\B{x}$ with style $\B{v}_{\B{x}}$ to style $\B{v}_{\B{y}}$ and
vice versa, i.e., to estimate the conditional distribution $p(\B{y} | \B{x})$
and $p(\B{x} | \B{y})$. Since text data is discrete, it is hard to learn the
transfer function directly via back-propagation as in computer
vision~\citep{zhu2017unpaired}. Instead, we assume the data is generated
conditioned on two disentangled parts, the style $\B{v}$ and the content
$\B{z}$\footnote{We drop the subscript in notations wherever the meaning is
  clear.}~\citep{hu2017toward}.

Consider the following generative process for each style: 1) the style representation
$\B{v}$ is sampled from a prior $p(\B{v})$; 2) the content vector $\B{z}$ is
sampled from $p(\B{z})$; 3) the sentence $\B{x}$ is generated from the
conditional distribution $p(\B{x}|\B{z}, \B{v})$. This model suggests the following parametric form for style transfer where $q$ represents a posterior:
\begin{align*}
  p(\B{y} | \B{x}) = \int_{\B{z}_{\B{x}}} p (\B{y}|\B{z}_{\B{x}}, \B{v}_{\B{y}}) q(\B{z}_{\B{x}} | \B{x}, \B{v}_{\B{x}}) d \B{z}_{\B{x}}.
\end{align*}
The above equation suggests the use of an encoder-decoder framework 
for style transfer problems. We can first encode the sentence $\B{x}$ to get
its content vector $\B{z}_{\B{x}}$, then we switch the style label from
$\B{v}_{\B{x}}$ to $\B{v}_{\B{y}}$. Combining the content vector $\B{z}_{\B{x}}$
and the style label $\B{v}_{\B{y}}$, we can generate a new sentence
$\tilde{\B{x}}$ (the transferred sentences are denotes as $\tilde{\B{x}}$ and
$\tilde{\B{y}}$).

One unsupervised approach is to use the auto-encoder model. We first use an
encoder model $\B{E}$ to encode $\B{x}$ and $\B{y}$ to get the content vectors $\B{z}_{\B{x}} = \B{E}(\B{x},
\B{v}_{\B{x}})$ and $\B{z}_{\B{y}} = \B{E}(\B{y}, \B{v}_{\B{y}})$. Then we use a
decoder $\B{G}$ to generate sentences conditioned on $\B{z}$ and $\B{v}$. The $\B{E}$ and $\B{G}$ together form an auto-encoder and the reconstruction loss is:
\begin{align*}
  \mathcal{L}_{\text{rec}}(\theta_{\B{E}}, \theta_{\B{G}})
  = \mathbb{E}_{\B{x}\sim\B{X}}[-\log p_{\B{G}}(\B{x}| \B{z}_{\B{x}}, \B{v}_{\B{x}})] 
  + \mathbb{E}_{\B{y}\sim\B{Y}}[-\log p_{\B{G}}(\B{y}| \B{z}_{\B{y}}, \B{v}_{\B{y}})],
\end{align*}
where $\B{v}_{\B{x}}$ and $\B{v}_{\B{y}}$ can be two learnable vectors to
represent the label embedding. In order to make sure that the $\B{z}_{\B{x}}$
and $\B{z}_{\B{y}}$ capture the content and we can deliver accurate transfer
between the style by switching the labels, we need to guarantee that
$\B{z}_{\B{x}}$ and $\B{z}_{\B{y}}$ follow the same distribution. We can
assume $p(\B{z})$ follows a prior distribution and add a KL-divergence
regularization on $\B{z}_{\B{x}}$, $\B{z}_{\B{y}}$. 
The model then becomes a VAE. However, previous
works~\citep{bowman2015generating,yang2017improved} found that there is a
training collapse problem with the VAE for text modeling and the posterior
distribution of $\B{z}$ fails to capture the content of a sentence.

To better capture the desired styles in the generated sentences, \cite{hu2017toward} additionally impose a style classifier on the generated samples, and the decoder $\B{G}$ is trained to generate sentences that maximize the accuracy of the style classifier. Such additional supervision with a \emph{discriminative} model is also adopted in \citep{shen2017style}, though in that work a binary real/fake classifier is instead used within a conventional adversarial scheme.

\paragraph{Adversarial Training}
\cite{shen2017style} use adversarial training to align the $\B{z}$
distributions. Not only do we want to align the distribution of $\B{z}_{\B{x}}$
and $\B{z}_{\B{y}}$, but also we hope that the transferred sentence
$\tilde{\B{x}}$ from $\B{x}$ to resemble $\B{y}$ and vice versa. 
Several adversarial discriminators are introduced to align these distributions. 
Each of the discriminators is a binary classifier distinguishing between real 
and fake. Specifically, the discriminator $D_{\B{z}}$ aims to distinguish 
between $\B{z}_{\B{x}}$ and $\B{z}_{\B{y}}$:
\begin{align*}
  \mathcal{L}_{\text{adv}}^{\B{z}}(\theta_{\B{E}}, \theta_{\B{D_{\B{z}}}})  = 
  \mathbb{E}_{\B{x}\sim\B{X}} [-\log D_{\B{z}}(\B{z}_{\B{x}})]
  + \mathbb{E}_{\B{y}\sim\B{Y}} [-\log (1 - D_{\B{z}}(\B{z}_{\B{y}}))] .
\end{align*}
Similarly, $D_{\B{x}}$ distinguish between $\B{x}$ and
$\tilde{\B{y}}$, yielding an objective $\mathcal{L}_{\text{adv}}^{\B{x}}$ as above; and $D_{\B{y}}$ distinguish between $\B{y}$ and $\tilde{\B{x}}$, yielding $\mathcal{L}_{\text{adv}}^{\B{y}}$. Since the samples
of $\tilde{\B{x}}$ and $\tilde{\B{y}}$ are discrete and it is hard to train
the generator in an end-to-end way, professor forcing~\citep{lamb2016professor} is used to match the distributions of the
hidden states of decoders. 
The overall training objective is a min-max game played among the encoder
$\B{E}$/decoder $\B{G}$ and the discriminators $D_{\B{z}}, D_{\B{x}}, D_{\B{y}}$~\citep{goodfellow2014generative}:
\begin{align*}
  \min_{E, G} \max_{D_{\B{z}}, D_{\B{x}}, D_{\B{y}}} \mathcal{L}_{\text{rec}} - \lambda( \mathcal{L}_{\text{adv}}^{\B{z}} + \mathcal{L}_{\text{adv}}^{\B{x}} + \mathcal{L}_{\text{adv}}^{\B{y}} )
\end{align*}
The model is trained in an alternating manner. In the first step, the loss of the
discriminators are minimize to distinguish between the $\B{z}_{\B{x}}, \B{x},
\B{y}$ and $\B{z}_{\B{y}}, \tilde{\B{x}}, \tilde{\B{y}}$, respectively; and in the
second step the encoder and decoder are trained to minimize the reconstruction
loss while maximizing loss of the discriminators.

\section{Language Models as Discriminators}\label{sec:lm}
In most past work, a classifier is used as the discriminator to distinguish whether a
sentence is real or fake. We propose instead to use locally-normalized language models as discriminators. We argue that using an explicit language model with token-level
locally normalized probabilities offers a more direct training signal to the
generator. If a transfered sentence does not match the target style, it will have high perplexity when
evaluated by a language model that was trained on target domain data. Not only does it
provide an overall evaluation score for the whole sentence, but a language model can
also assign a probability to each token, thus providing more information on which word
is to blame if the overall perplexity is very high.

\begin{figure}[!t]
  \centering \includegraphics[width=0.6\textwidth]{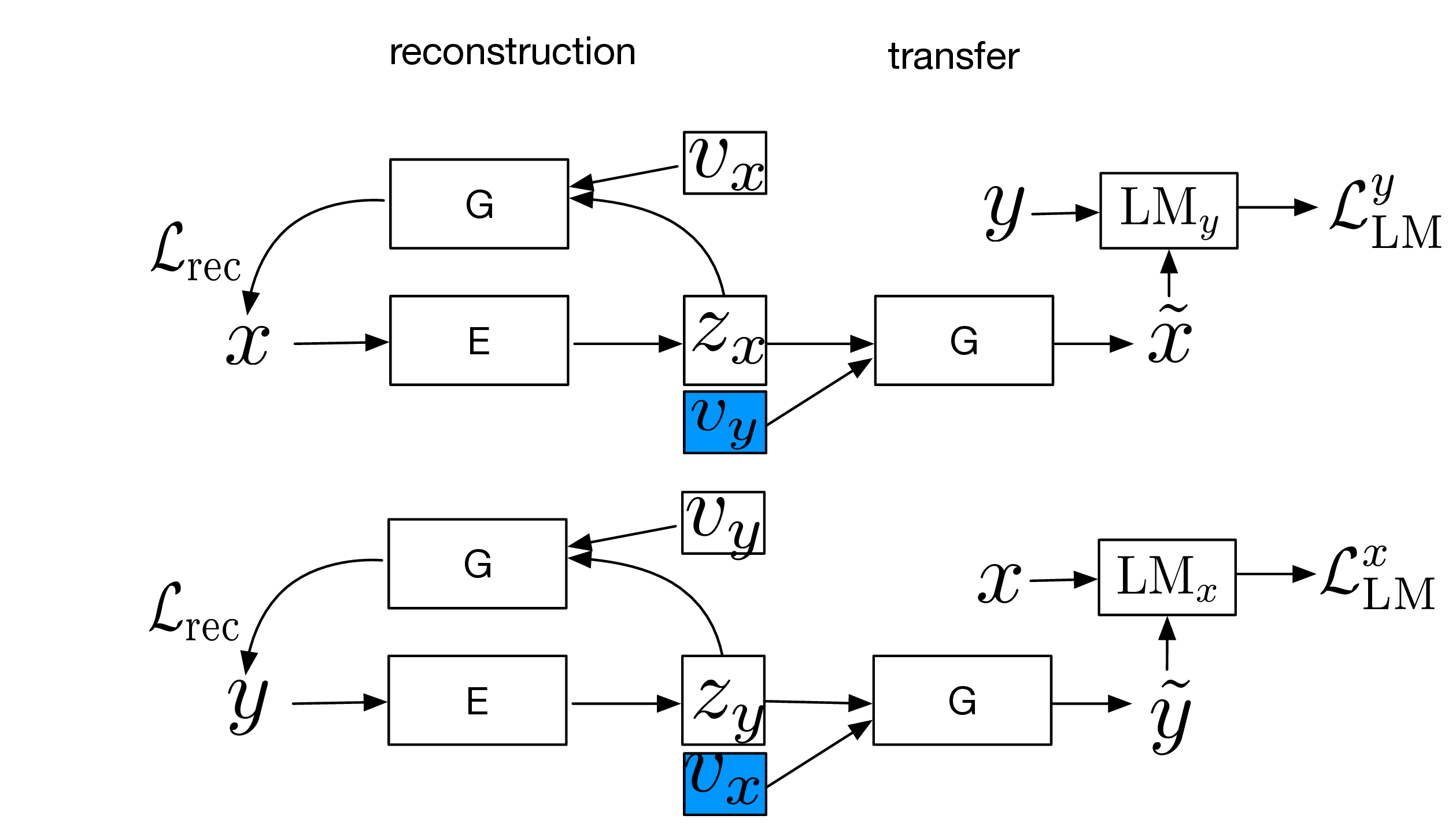}
  \captionof{figure}{The overall model architecture consists of two parts:
    reconstruction and transfer. For transfer, we switch the style label and sample an output sentence from the generator that is evaluated by a language model.} \label{fig:model}
    \vspace{-0.3cm}
\end{figure}

The overall model architecture is shown in Figure~\ref{fig:model}.
Suppose $\tilde{\B{x}}$ is the output sentence from applying style transfer to input sentence
$\B{x}$, i.e., $\tilde{\B{x}}$ is sampled from
$p_G(\tilde{\B{x}}| \B{z}_{\B{x}}, \B{v}_{\B{y}})$ (and similary for $\tilde{\B{y}}$ and $\B{y}$). Let $p_{\text{LM}}(\B{x})$ be the probability
of a sentence $\B{x}$ evaluate against a language model, then the discriminator
loss becomes:
\begin{align}
  \mathcal{L}_{\text{LM}}^{\B{x}}(\theta_{\B{E}}, \theta_{\B{G}}, \theta_{\B{\text{LM}_{\B{x}}}})
  & = \mathbb{E}_{\B{x}\sim\B{X}} [-\log p_{\text{LM}_{\B{x}}}(\B{x}))] 
    + \gamma \mathbb{E}_{\B{y}\sim\B{Y}, \tilde{\B{y}} \sim p_G(\tilde{\B{y}}| \B{z}_{\B{y}}, \B{v}_{\B{x}}
    ) } [\log p_{\text{LM}_{\B{x}}}(\tilde{\B{y}})], \label{eq:lm1} \\ 
  \mathcal{L}_{\text{LM}}^{\B{y}}(\theta_{\B{E}}, \theta_{\B{G}}, \theta_{\B{\text{LM}_{\B{y}}}})
  & = \mathbb{E}_{\B{y}\sim\B{Y}} [-\log p_{\text{LM}_{\B{y}}}(\B{y}))] 
    + \gamma \mathbb{E}_{\B{x}\sim\B{X}, \tilde{\B{x}} \sim p_G(\tilde{\B{x}} | \B{z}_{\B{x}}, \B{v}_{\B{y}}) } [\log p_{\text{LM}_{\B{y}}}(\tilde{\B{x}})] . \label{eq:lm2}
\end{align}

Our overall objective becomes:
\begin{align}
  \min_{E, G} \max_{\text{LM}_{\B{x}}, \text{LM}_{\B{y}}} \mathcal{L}_{\text{rec}} - \lambda(\mathcal{L}_{\text{LM}}^{\B{x}} + \mathcal{L}_{\text{LM}}^{\B{y}} ) \label{eq:lm_overall}
\end{align}

\textbf{Negative samples}: Note that Equation~\ref{eq:lm1} and \ref{eq:lm2}
differs from traditional ways of training language models in that we have a term
including the negative samples. We train the LM in an adversarial way by
minimizing the loss of LM of real sentences and maximizing the loss of
transferred sentences. However, since the LM is a structured discriminator, we would
hope that a language model trained on the real sentences will automatically
assign high perplexity to sentences not in the target domain, hence negative
samples from the generator may not be necessary. To investigate the necessity of
negative samples, we add a weight $\gamma$ to the loss of negative samples.
The weight $\gamma$ adjusts the negative sample loss in training
the language models. If $\gamma=0$, we simply train the language model on
real sentences and fix its parameters, avoiding potentially unstable adversarial training steps. We investigate the necessity of using negative samples in the
experiment section.

Training consists of two steps alternatively. In the first step, we train
the language models according to Equation~\ref{eq:lm1} and~\ref{eq:lm2}. In the
second step, we minimize the reconstruction loss as well as the perplexity
of generated samples evaluated by the language model. Since $\tilde{\B{x}}$ is
discrete, one can use the REINFORCE~\citep{sutton2000policy} algorithm to train
the generator:
\begin{align}
  \nabla_{\theta_G} \mathcal{L}_{\text{LM}}^{\B{y}} = \mathbb{E}_{\B{x}\sim\B{X}, \tilde{\B{x}} \sim p_G(\tilde{\B{x}}|\B{z}_{\B{x}},
  \B{v}_{\B{y}})}[\log p_{\text{LM}}(\tilde{\B{x}})\nabla_{\theta_G}  \log p_G(\tilde{\B{x}}|\B{z}_{\B{x}}, \B{v}_{\B{y}})] \label{eq:reinforce}.
\end{align}
However, using a single sample to approximate the expected gradient
leads to high variance in gradient estimates and thus unstable learning.

\begin{figure}[!t]
  \centering \includegraphics[width=0.6\textwidth]{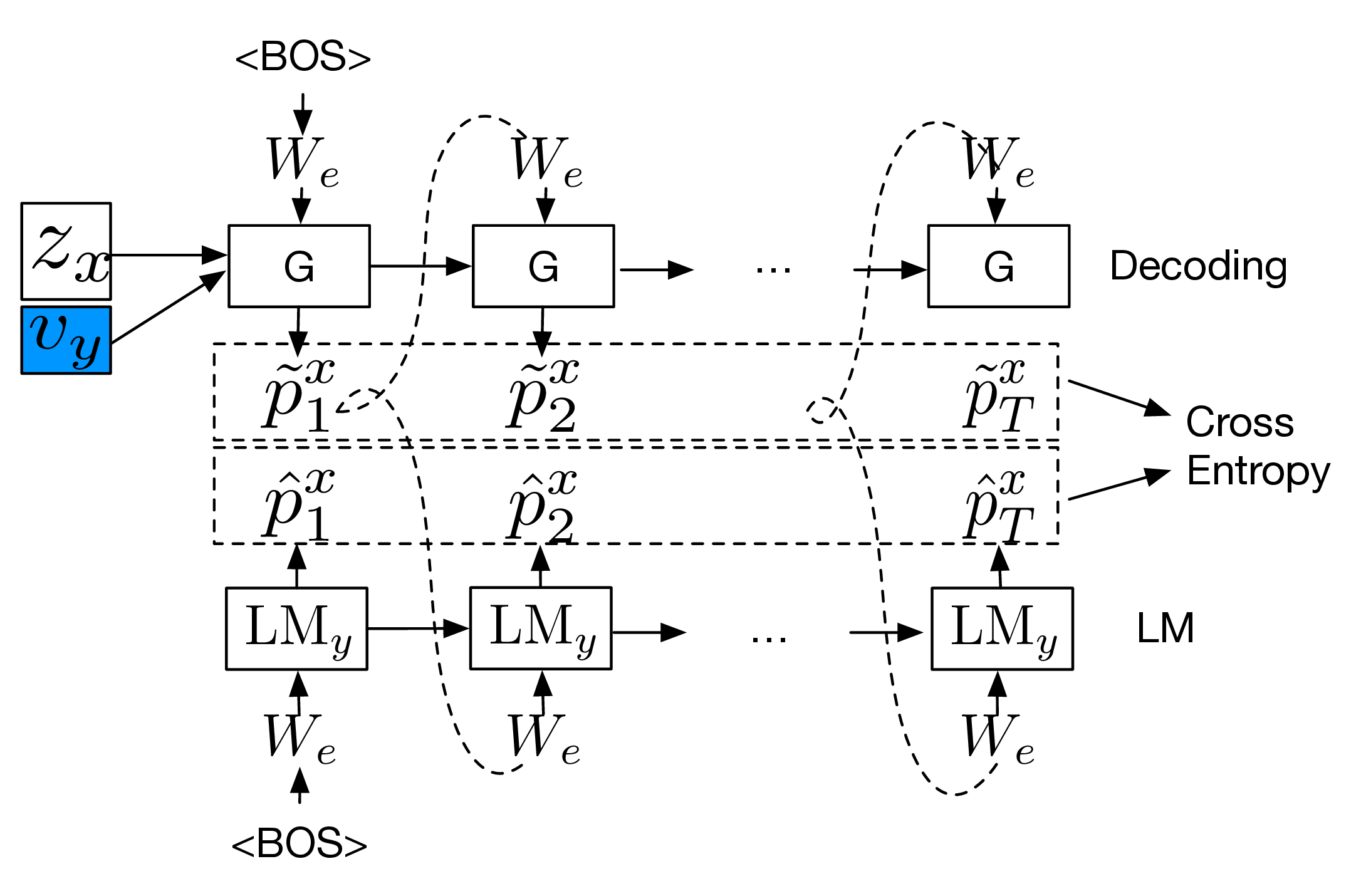}
  \captionof{figure}{Continuous approximation of language model loss. The input
    is a sequence of probability distributions $\{\tilde{\B{p}}_t^{\B{x}}\}_{t=1}^T$ sampled
    from the generator. At each timestep, we compute a weighted embedding as input to the language
    model and get the sequence of output distributions from the LM as
    $\{\hat{\B{p}}_t^{\B{x}}\}_{t=1}^T$. The loss is the sum of cross entropies between each pair of
    $\tilde{\B{p}}_t^{\B{x}}$ and $\hat{\B{p}}_t^{\B{x}}$.}
  \label{fig:lm}
\end{figure}

{\bf Continuous approximation}: Instead, we propose to use a continuous approximation to the sampling process in training the generator, as demonstrated in Figure~\ref{fig:lm}.
Instead of feeding a
single sampled word as input to the next timestep of the generator, we use a
Gumbel-softmax~\citep{jang2016categorical} distribution as a continuous
approximation to sample instead. Let $u$ be a categorical distribution with
probabilities $\pi_1, \pi_2, \ldots, \pi_c$. Samples from $u$ can be
approximated using:
\begin{align*}
  p_i = \frac{\exp((\log \pi_i) + g_i)/\tau}{\sum_{j=1}^c \exp((\log \pi_j + g_j)/\tau)},
\end{align*}
where the $g_i$'s are independent samples from $\mathrm{Gumbel}(0, 1)$.

Let the tokens of the transferred sentence be $\tilde{\B{x}} = \{ \tilde{x}_t
\}_{t=1}^T$. Suppose the output of the logit at timestep $t$ is $\B{v}_t^{\B{x}}$,
then $\tilde{{\B{p}}}_t^{\B{x}}= \text{Gumbel-softmax}(\B{v}_t^{\B{x}}, \tau)$,
where $\tau$ is the temperature. When $\tau \to 0$, $\tilde{\B{p}}_t^{\B{x}}$
becomes the one hot representation of token $\tilde{x}_t$. Using the continuous
approximation, then the output of the decoder becomes a sequence of probability
vectors $\tilde{\B{p}}^{\B{x}} = \{ \tilde{\B{p}}_t^{\B{x}}\}_{t=1}^{T}$.

With the continuous approximation of $\tilde{\B{x}}$, we can calculate the loss
evaluated using a language model easily, as shown in Figure~\ref{fig:lm}. For
every step, we feed $\tilde{\B{p}}_t^{\B{x}}$ to the language model of $\B{y}$
(denoted as $\text{LM}_{\B{y}}$) using the weighted average of the embedding
$W_e\tilde{\B{p}}_t^{\B{x}}$, then we get the output from the
$\text{LM}_{\B{y}}$ which is a probability distribution over the vocabulary of
the next word $\hat{\B{p}}_{t+1}^{\B{x}}$. The loss of the current step is the
cross entropy loss between $\tilde{\B{p}}_{t+1}^{\B{x}}$ and
$\hat{\B{p}}_{t+1}^{\B{x}}$: $(\tilde{\B{p}}_{t+1}^{\B{x}})^\intercal\log
\hat{\B{p}}_{t+1}^{\B{x}} $. Note that when the decoder output distribution
$\tilde{\B{p}}_{t+1}^{\B{x}}$ aligns with the language model output distribution
$\hat{\B{p}}_{t+1}^{\B{x}}$, the above loss achieves minimum. By summing the
loss over all steps and taking the gradient, we can use standard
back-propagation to train the generator:
\begin{align}
  \nabla_{\theta_G}{\mathcal{L}}_{\text{LM}}^{\B{y}} \approx \mathbb{E}_{\B{x}\sim\B{X}, \tilde{\B{p}}^{\B{x}} \sim p_G(\tilde{\B{x}}|\B{z}_{\B{x}},
  \B{v}_{\B{y}})} [\nabla_{\theta_G} \sum_{t=1}^T (\tilde{\B{p}}_t^{\B{x}})^\intercal\log \hat{\B{p}}_t^{\B{x}}] \label{eq:gradient}.
\end{align}
The above Equation is a continuous approximation of Equation~\ref{eq:reinforce}
with Gumbel softmax distribution. In experiments, we use a single sample of
$\tilde{\B{p}}^{\B{x}}$ to approximate the expectation.

Note that the use of the language model discriminator is a somewhat different in
each of the two types of training update steps because of the continuous approximation. We use discrete
samples from the generators as negative samples in training the language model
discriminator step, while we use a continuous approximation in updating the
generator step according to Equation~\ref{eq:gradient}.


\textbf{Overcoming mode collapse}: It is known that in adversarial training, the
generator can suffer from mode collapse~\citep{arjovsky2017towards,hu2017unifying} where the
samples from the generator only cover part of the data distribution. In
preliminary experimentation, we found that the language model prefers short
sentences. To overcome this length bias, we use two tricks in our experiments:
1) we normalize the loss of Equation~\ref{eq:gradient} by length and 2) we fix
the length of $\tilde{\B{x}}$ to be the same of $\B{x}$.
We find these two tricks stabilize the training and avoid generating collapsed
overly short outputs.

\section{Experiments}
In order to verify the effectiveness of our model, we experiment on three tasks:
word substitution decipherment, sentiment modification, and related language
translation. We mainly compare with the most comparable approach of \citep{shen2017style} that
uses CNN classifiers as discriminators\footnote{We use the code from 
\url{https://github.com/shentianxiao/language-style-transfer}. }. Note that~\cite{shen2017style} use
three discriminators to align both $\B{z}$ and decoder hidden states, while
our model only uses a single language model as a discriminator directly on the output sentences
$\tilde{\B{x}}, \tilde{\B{y}}$. Moreover, we also compare with a broader set of related work~\citep{hu2017toward,fu2017style,li2018delete} for the tasks when appropriate. Our proposed model provides substantiate improvements in most of the cases.
We implement our model with the Texar~\citep{hu2018texar} toolbox based on
Tensorflow~\citep{abadi2016tensorflow}.

\subsection{Word substitution decipherment}
As the first task, we consider the word substitution decipherment task previous
explored in the NLP literature~\citep{dou2012large}. We can control the amount of
change to the original sentences in word substitution decipherment so as to
systematically investigate how well the language model performs in a task that
requires various amount of changes. In word substitution cipher, every token in the
vocabulary is mapped to a cipher token and the tokens in sentences are replaced
with cipher tokens according to the cipher dictionary. The task of decipherment is
to recover the original text without any knowledge of the dictionary.

\textbf{Data}: Following~\citep{shen2017style}, we sample 200K sentences from the
Yelp review dataset as plain text $\B{X}$ and sample other 200K sentences and
apply word substitution cipher on these sentences to get $\B{Y}$. We use another
100k {\em parallel} sentences as the development and test set respectively.
Sentences of length more than 15 are filtered out. We keep all words that appear
more than 5 times in the training set and get a vocabulary size of about 10k.
All words appearing less than 5 times are replaced with a “<unk>” token. We random
sample words from the vocabulary and replace them with cipher tokens. The amount
of ciphered words ranges from 20\% to 100\%. 
As we have ground truth plain text, we can directly measure the 
BLEU~\footnote{BLEU score is measured with \texttt{multi-bleu.perl}.}
score to evaluate the model. Our model configurations are included in 
Appendix~\ref{sec:config}.

\begin{table}[!t]
    \centering
	\begin{tabular}{@{}l r r r r r@{}}
    \toprule
    Model & 20\% & 40\% & 60\% & 80\% & 100\% \\
    \midrule
    Copy & 64.3 &	39.1	& 14.4 & 2.5 &	0 \\
    \cite{shen2017style}$^*$ & 86.6 & 77.1 & 70.1 & 61.2 & \textbf{50.8} \\
    \midrule
    \texttt{Our results:} \\
    LM & 89.0 & \textbf{80.0} & \textbf{74.1} & 62.9 & 49.3 \\
    LM + adv & \textbf{89.1} & 79.6 & 71.8 & \textbf{63.8} & 44.2 \\
    \cmidrule[\heavyrulewidth]{1-6}
    \end{tabular}
	  \caption{Decipherment results measured in BLEU. Copy is directly measuring
      $\B{y}$ against $\B{x}$. LM + adv denotes we use negative samples to
      train the language model.$^*$We run the code open-sourced by the authors
      to get the results.} \label{tab:decipher}
\end{table}
\begin{table}
\centering
    \begin{tabular}{@{}l r r r r@{}}
      \toprule
      Model & Accu & BLEU & $\text{PPL}_{\B{X}}$ & $\text{PPL}_{\B{Y}}$ \\
      \midrule
      \cite{shen2017style} & 79.5 & 12.4 & 50.4 & 52.7 \\
      \cite{hu2017toward} & 87.7 & {\bf 65.6} & 115.6 & 239.8 \\
      \midrule
      \texttt{Our results:} \\
      LM & 83.3 & 38.6 & {\bf 30.3} & {\bf 42.1} \\
      LM + Classifier & {\bf 91.2} & 57.8 & 47.0 & 60.9 \\
      \cmidrule[\heavyrulewidth]{1-5}
    \end{tabular}
    \caption{Results for sentiment modification. $\B{X} = \text{negative}, \B{Y}
      = \text{positive}$. $\text{PPL}_{\B{x}}$ denotes the perplexity of
      sentences transferred from positive sentences evaluated by a language
      model trained with negative sentences and vice versa. }
  \label{tab:tsf}
  \vspace{-0.3cm}
\end{table}

\textbf{Results}: The results are shown in Table~\ref{tab:decipher}. We first
investigate the effect of using negative samples in training the language model,
as denotes by LM + adv in Table~\ref{tab:decipher}. We can see that using
adversarial training sometimes improves the results. However, we found
empirically that using negative samples makes the training very unstable and the
model diverges easily. This is the main reason why we did not get consistently
better results by incorporating adversarial training.

Comparing with ~\citep{shen2017style}, we can see that the language model
without adversarial training is already very effective and performs much better
when the amount of change is less than 100\%. This is intuitive because when the
change is less than 100\%, a language model can use context information to
predict and correct enciphered tokens. It's surprising that even with 100\%
token change, our model is only 1.5 BLEU score worse than
~\citep{shen2017style}, when all tokens are replaced and no context information
can be used by the language model. We guess our model can gradually decipher
tokens from the beginning of a sentence and then use them as a bootstrap to
decipher the whole sentence. We can also combine language models with the CNNs 
as discriminators. For example,
for the 100\% case, we get BLEU score of 52.1 when combing them.
Given unstableness of adversarial training and effectiveness
of language models, we set $\gamma=0$ in Equation~\ref{eq:lm1} and~\ref{eq:lm2}
in the rest of the experiments.

\subsection{Sentiment Manipulation}
We have demonstrated that the language model can successfully
crack word substitution cipher. However, the change of substitution cipher is limited to
a one-to-one mapping. As the second task, we would like to investigate whether a
language model can distinguish sentences with positive and negative sentiments,
thus help to transfer the sentiments of sentences while preserving the content.
We compare to the model of ~\citep{hu2017toward} as an additional baseline,
which uses a pre-trained classifier as guidance.

\textbf{Data}: We use the same data set as in ~\citep{shen2017style}. The data
set contains 250K negative sentences (denoted as $\B{X}$) and 380K positive
sentences (denoted as $\B{Y}$), of which 70\% are used for training, 10\% are used
for development and the remaining 20\% are used as test set. The pre-processing
steps are the same as the previous experiment. We also use similar experiment
configurations.

\textbf{Evaluation}:
Evaluating the quality of transferred sentences is a challenging problem as
there are no ground truth sentences. We follow previous papers in using
model-based evaluation. We measure whether transferred sentences have the
correct sentiment according to a pre-trained sentiment classifier. We follow
both \citep{hu2017toward} and \citep{shen2017style} in using a CNN-based
classifier. However, simply evaluating the
sentiment of sentences is not enough since the model can output collapsed output
such as a single word ``good'' for all negative transfer and ``bad'' for all
positive transfer. We not only would like transferred sentences to preserve the content
of original sentences, but also to be smooth in terms of language quality.
For these two aspects, we propose to measure the BLEU score of transferred sentences against original
sentences and measure the perplexity of transferred sentences to evaluate the
fluency. A good model should perform well on all three metrics.

\textbf{Results}: We report the results in Table.~\ref{tab:tsf}. 
As a baseline, the original corpus has perplexity of $35.8$ and $38.8$ for the
negative and positive sentences respectively. Comparing LM with
~\citep{shen2017style}, we can see that LM outperforms it in all three aspects:
getting higher accuracy, preserving the content better while being more fluent.
This demonstrates the effectiveness of using LM as the discriminator.
\citep{hu2017toward} has the highest accuracy and BLEU score among the three
models while the perplexity is very high. It is not surprising that the
classifier will only modify the features of the sentences that are related to the
sentiment and there is no mechanism to ensure that the modified sentence being
fluent. Hence the corresponding perplexity is very high. We can manifest the
best of both models by combing the loss of LM and the classifier in
\citep{hu2017toward}: a classifier is good at modifying the sentiment and an LM
can smooth the modification to get a fluent sentence. We find improvement of
accuracy and perplexity as denoted by LM + classifier compared to
classifier only \citep{hu2017toward}.

\textbf{Comparing with other models}: Recently there are other models that are
proposed specifically targeting the sentiment modification task such as
~\citep{li2018delete}. Their method is feature based and consists of the
following steps: (\texttt{Delete}) first, they use the statistics of word
frequency to delete the attribute words such as ``good, bad'' from original
sentences, (\texttt{Retrieve}) then they retrieve the most similar sentences
from the other corpus based on nearest neighbor search, (\texttt{Generate}) the
attribute words from retrieved sentences are combined with the content words of
original sentences to generate transferred sentences. The authors provide 500
human annotated sentences as the ground truth of transferred sentences so we measure
the BLEU score against those sentences. The results are shown in
Table~\ref{tab:stanford}. We can see our model has similar accuracy compared
with DeleteAndRetrieve, but has much better BLEU scores and slightly better 
perplexity.

We list some examples of transferred sentences in Table~\ref{tab:examples} in
the appendix. We can see that ~\citep{shen2017style} does not keep the content
of the original sentences well and changes the meaning of the original
sentences. \citep{hu2017toward} changes the sentiment but uses improper words,
e.g. ``maintenance is equally \texttt{hilarious}''. Our LM can change the
change the sentiment of sentences. But sometimes there is an over-smoothing
problem, changing the less frequent words to more frequent words, e.g. changing
``my goodness it was so gross'' to ``my \texttt{food} it was so good.''. In
general LM + classifier has the best results, it changes the sentiment, while
keeps the content and the sentences are fluent.

\begin{table}
\centering
	\begin{tabular}{@{}l r r r r@{}}
    \toprule
    Model & ACCU & BLEU	& $\text{PPL}_{\B{X}}$ & $\text{PPL}_{\B{Y}}$ \\
    \cite{shen2017style} & 76.2 &	6.8	& 49.4 & 45.6 \\
    \cite{fu2017style}: \\
    StyleEmbedding & 	9.2 &	16.65 &	97.51	& 142.6 \\
    MultiDecoder	& 50.9 & 11.24 & 111.1 & 119.1 \\
    \midrule
    \cite{li2018delete}: \\
    Delete	& 87.2 & 11.5 &	75.2 & 68.7 \\
    Template & 86.7 &	18.0 & 192.5 & 148.4 \\
    Retrieval	& {\bf 95.1} & 1.3 & {\bf 31.5} & {\bf 37.0} \\
    DeleteAndRetrieval & 90.9 & 12.6 & 104.6 & 43.8 \\
    \midrule
    {\bf \texttt{Our results:}} \\
    LM & 85.4 & 13.4 & 32.8	& 40.5 \\
    LM + Classifier & 90.0 & {\bf 22.3} & 48.4 & 61.6 \\
    \cmidrule[\heavyrulewidth]{1-5}
    \end{tabular}
    \caption{Results for sentiment modification based on the 500 human annotated
      sentences as ground truth from ~\citep{li2018delete}.} \label{tab:stanford}
      \vspace{-0.2cm}
\end{table}

\subsection{Related language translation}
In the final experiment, we consider a more challenging task: unsupervised
related language translation~\citep{pourdamghani2017deciphering}. Related
language translation is easier than normal pair language translation since there
is a close relationship between the two languages. Note here we don't compare with
other sophisticated unsupervised neural machine translation systems such as
~\citep{lample2017unsupervised, artetxe2017unsupervised}, whose models are much
more complicated and use other techniques such as back-translation, but simply
compare the different type of discriminators in the context of a simple model.

\textbf{Data}: We choose Bosnian (bs) vs Serbian (sr) and simplified Chinese
(zh-CN) vs traditional Chinese (zh-TW) pair as our experiment languages. Due to
the lack of parallel data for these data, we build the data ourselves. For bs
and sr pair, we use the monolingual data from Leipzig Corpora
Collections\footnote{\url{http://wortschatz.uni-leipzig.de/en}}.
We use the news data and sample about 200k sentences of length less than 20 for
each language, of which 80\% are used for training, 10\% are used for validation
and remaining 10\% are used for test. For validation and test, we obtain the
parallel corpus by using the Google Translation
API
.
The vocabulary size is 25k for the sr vs bs language pair. For zh-CN and zh-TW
pair, we use the monolingual data from the Chinese Gigaword corpus. We use the
news headlines as our training data. 300k sentences are sampled for each
language. The data is partitioned and parallel data is obtained in a similar way to
that of sr vs bs pair. We directly use a character-based model and the total
vocabulary size is about 5k. For evaluation, we directly measure the BLEU score
using the references for both language pairs.

Note that the relationship between zh-CN and zh-TW is simple and mostly like a
decipherment problem in which some simplified Chinese characters have the
corresponding traditional character mapping. The relation between bs vs sr is
more complicated.

\textbf{Results}: The results are shown in Table.~\ref{tab:translation}. For
sr--bos and bos--sr, since the vocabulary of two languages does not overlap at
all, it is a very challenging task. We report the BLEU1 metric since the BLEU4
is close to 0. We can see that our language model discriminator still
outperforms \citep{shen2017style} slightly. The case for zh--tw and tw--zh is
much easier. Simple copying already has a reasonable score of 32.3. Using our
model, we can improve it to 81.6 for cn--tw and 85.5 for tw--cn, outperforming
\citep{shen2017style} by a large margin.

\section{Related Work}
\textbf{Non-parallel transfer in natural language}: \citep{hu2017toward,
  shen2017style, prabhumoye2018style, gomez2018unsupervised} 
are most relevant to our work. \cite{hu2017toward} aim to
generate sentences with controllable attributes by learning disentangled
representations. \cite{shen2017style} introduce adversarial training to
unsupervised text style transfer. They apply discriminators both on the encoder
representation and on the hidden states of the decoders to ensure that they have
the same distribution. These are the two models that we mainly compare with.
\cite{prabhumoye2018style} use the back-translation technique in their model, which
is complementary to our method and can be integrated into our model to further
improve performance. \cite{gomez2018unsupervised} use GAN-based approach
to decipher shift ciphers.
\citep{lample2017unsupervised, artetxe2017unsupervised} propose unsupervised
machine translation and use adversarial training to match the encoder
representation of the sentences from different languages. They also use
back-translation to refine their model in an iterative way.

\begin{table}
\centering
    \begin{tabular}{@{}l r r r r@{}}
      \toprule
      Model & sr--bs & bs--sr & cn--tw & tw--cn \\
      \midrule
      Copy & 0 & 0	& 32.3 & 32.3 \\
      \cite{shen2017style} & 29.1 & 30.3 & 60.1 & 60.7 \\
      \midrule
      \texttt{Our results:} \\
      LM & \textbf{31.0} & \textbf{31.7} & \textbf{81.6} & \textbf{85.5} \\
      \cmidrule[\heavyrulewidth]{1-5}
    \end{tabular}
    \caption{Related language translation results measured in BLEU. The results
      for sr vs bs in measured in BLEU1 while cn vs tw is measure in BLEU.
    } \label{tab:translation}
\end{table}

\textbf{GANs}: GANs have been widely explored recently, especially in computer
vision~\citep{zhu2017unpaired, chen2016infogan, radford2015unsupervised,
  sutton2000policy, salimans2016improved, denton2015deep, isola2017image}. The
progress of GANs on text is relatively limited due to the non-differentiable
discrete tokens. Lots of papers~\citep{yu2017seqgan, che2017maximum,
  li2017adversarial, yang2017improving} use REINFORCE~\citep{sutton2000policy}
to finetune a trained model to improve the quality of samples. There is also
prior work that attempts to introduce more structured discriminators, for
instance, the energy-based GAN (EBGAN)~\citep{zhao2016energy}
and RankGAN~\citep{lin2017adversarial}. 
Our language model can be seen as a special energy function, but it is more
complicated than the auto-encoder used in \citep{zhao2016energy} since it has a
recurrent structure. \cite{hu2018deep} also proposes to use structured discriminators
in generative models and establishes its the connection with posterior 
regularization.

\textbf{Computer vision style transfer}: Our work is also related to
unsupervised style transfer in computer vision~\citep{gatys2016image,
  huang2017arbitrary}. 
\citep{gatys2016image} directly uses the covariance matrix of the CNN features
and tries to align the covariance matrix to transfer the style.
\citep{huang2017arbitrary} proposes adaptive instance normalization for an arbitrary
style of images. \citep{zhu2017unpaired} uses a cycle-consistency loss to ensure
the content of the images is preserved and can be translated back to original
images.

\textbf{Language model for reranking}: Previously, language models are used to
incorporate the knowledge of monolingual data mainly by reranking the sentences
generated from a base model such as ~\citep{brants2007large, gulcehre2015using,he2016dual}. \citep{liu2017unsupervised,chen2016unsupervised}
use a language model as training supervision for unsupervised OCR. 
Our model is more advanced in using language models as
discriminators in distilling the knowledge of monolingual data to a base model
in an end-to-end way. 

\section{Conclusion}
We showed that by using language models as discriminators and we could outperform
traditional binary classifier discriminators in three unsupervised text style
transfer tasks including word substitution decipherment, sentiment modification
and related language translation. In comparison with a binary classifier
discriminator, a language model can provide a more stable and more informative
training signal for training generators. Moreover, we empirically found that 
it is possible to eliminate adversarial training with negative samples if a structured
model is used as the discriminator, thus pointing one possible direction to solve 
the training difficulty of GANs. In the future, we plan to explore and extend our 
model to semi-supervised learning.

\bibliographystyle{abbrvnat}
\bibliography{bibfile}

\appendix
\newpage
\section{Training Algorithms}
\begin{algorithm}[!h]
  \begin{algorithmic}[0]
    \Input Data set of two different styles $\B{X}$, $\B{Y}$. \State Parameters:
    weight $\lambda$ and $\gamma$, temperature $\tau$. \State Initialized model
    parameters $\theta_\B{E}, \theta_\B{G}, \theta_{\text{LM}_{\B{x}}},
    \theta_{\text{LM}_{\B{y}}}$. \Repeat \State Update
    $\theta_{\text{LM}_{\B{x}}}$ and $\theta_{\text{LM}_{\B{y}}}$ by minimizing
    $\mathcal{L}_{\text{LM}}^{\B{x}}(\theta_{\text{LM}_{\B{x}}})$ and
    $\mathcal{L}_{\text{LM}}^{\B{y}}(\theta_{\text{LM}_{\B{y}}})$ respectively.
    \State Update $\theta_{\B{E}}, \theta_{\B{G}}$ by minimizing:
    $\mathcal{L}_{\text{rec}} - \lambda (\mathcal{L}^{\B{x}}_{\text{LM}} +
    \mathcal{L}^{\B{y}}_{\text{LM}}) $ using Equation~\ref{eq:gradient}.
    \Until{convergence} \Output A text style transfer model with parameters
    $\theta_{\B{E}}$, $\theta_{\B{G}}$.
      \end{algorithmic}
      \caption{Unsupervised text style transfer.}
      \label{alg:tsf}
    \end{algorithm}

\section{Model Configurations} \label{sec:config}
Similar model configuration to that of ~\citep{shen2017style} is used for a fair
comparison. We use one-layer GRU~\citep{chung2014empirical} as the encoder and
decoder (generator). We set the word embedding size to be $100$ and GRU hidden
size to be $700$. $\B{v}$ is a vector of size $200$. For the language model, we
use the same architecture as the decoder. The parameters of the language model
are not shared with parameters of other parts and are trained from scratch. We
use a batch size of $128$, which contains 64 samples from $\B{X}$ and $\B{Y}$
respectively. We use Adam~\citep{kingma2014adam} optimization algorithm to train
both the language model and the auto-encoder and the learning rate is set to be
the same. Hyper-parameters are selected based on the validation set. We use grid
search to pick the best parameters. The learning rate is selected from $[1e-3,
5e-4, 2e-4, 1e-4]$ and $\lambda$, the weight of language model loss, is selected
from $[1.0, 0.5, 0.1]$. Models are trained for a total of 20 epochs. We use an
annealing strategy to set the temperature of $\tau$ of the Gumbel-softmax
approximation. The initial value of $\tau$ is set to 1.0 and it decays by half
every epoch until reaching the minimum value of 0.001.

\newpage
\section{Sentiment Transfer Examples}
\begin{table}[!b]
  \small
  \begin{center}
    \begin{tabular}{l l l}
      \toprule
      Model & Negative to Positive \\
      \midrule
      Original  & it was super dry and had a weird taste to the entire slice .\\
      \citep{shen2017style} & it was super friendly and had a nice touch to the same . \\
      \citep{hu2017toward} & it was super well-made and had a weird taste to the entire slice . \\
      LM & it was very good , had a good taste to the food service . \\
      LM + classifier & it was super fresh and had a delicious taste to the entire slice . \\
      \newline \\  
      Original & my goodness it was so gross . \\
      \citep{shen2017style} & my server it was so . \\
      \citep{hu2017toward} & my goodness it was so refreshing . \\
      LM  & my food it was so good . \\
      LM + classifier & my goodness it was so great . \\
      \newline \\
      Original & maintenance is equally incompetent . \\
      \citep{shen2017style} & everything is terrific professional . \\
      \citep{hu2017toward} & maintenance is equally hilarious . \\
      LM  & maintenance is very great . \\
      LM + classifier & maintenance is equally great . \\
      \newline \\
      Original & if i could give them a zero star review i would ! \\
      \citep{shen2017style} & if i will give them a breakfast star here ever ! \\
      \citep{hu2017toward} & if i lite give them a sweetheart star review i would ! \\
      LM & if i could give them a \_num\_ star place i would . \\
      LM + classifier & if i can give them a great star review i would ! \\
      \bottomrule
      \newline \\
      \toprule
      Model & Positive to Negative \\
      \midrule
      Original & did n't know this type cuisine could be this great ! \\
      \citep{shen2017style} & did n't know this old food you make this same horrible ! \\
      \citep{hu2017toward} & did n't know this type cuisine could be this great ! \\
      LM & did n't know this type , could be this bad .\\
      LM + classifier & did n't know this type cuisine could be this horrible .\\
      \newline \\
      Original & besides that , the wine selection they have is pretty awesome as well .\\
      \citep{shen2017style} & after that , the quality prices that does n't pretty much well as .\\
      \citep{hu2017toward} & besides that , the wine selection they have is pretty borderline as atrocious . \\
      LM & besides that , the food selection they have is pretty awful as well . \\
      LM + classifier & besides that , the wine selection they have is pretty horrible as well . \\
      \newline \\
      Original & uncle george is very friendly to each guest . \\
      \citep{shen2017style} & if there is very rude to our cab . \\
      \citep{hu2017toward} & uncle george is very lackluster to each guest . \\
      LM & uncle george is very rude to each guest . \\
      LM + classifier & uncle george is very rude to each guest . \\
      \newline \\
      Original & the food is fresh and the environment is good . \\
      \citep{shen2017style} & the food is bland and the food is the nightmare . \\
      \citep{hu2017toward} & the food is atrocious and the environment is atrocious . \\
      LM & the food is bad , the food is bad . \\
      LM + classifier & the food is bland and the environment is bad .\\
      \bottomrule
    \end{tabular}
    \caption{Sentiment transfer examples.}
  \label{tab:examples}
\end{center}
\end{table}

\end{document}